# WEDepth: Efficient Adaptation of World Knowledge for Monocular Depth Estimation

Gongshu Wang[*], Zhirui Wang[*], Kan Yang[*]

* Aerospace Information Research Institute, Chinese Academy of Sciences

## Abstract

Monocular depth estimation (MDE) has widely applicable but remains highly challenging due to the inherently ill-posed nature of reconstructing 3D scenes from single 2D images. Modern Vision Foundation Models (VFMs), pre-trained on large-scale diverse datasets, exhibit remarkable world understanding capabilities that benefit for various vision tasks. Recent studies have demonstrated significant improvements in MDE through fine-tuning these VFMs. Inspired by these developments, we propose WEDepth, a novel approach that adapts VFMs for MDE without modifying their structures and pretrained weights, while effectively eliciting and leveraging their inherent priors. Our method employs the VFM as a multi-level feature enhancer, systematically injecting prior knowledge at different representation levels. Experiments on NYU-Depth v2 and KITTI datasets show that WEDepth establishes new state-of-the-art (SOTA) performance, achieving competitive results compared to both diffusion-based approaches (which require multiple forward passes) and methods pretrained on relative depth. Furthermore, we demonstrate our method exhibits strong zero-shot transfer capability across diverse scenarios.

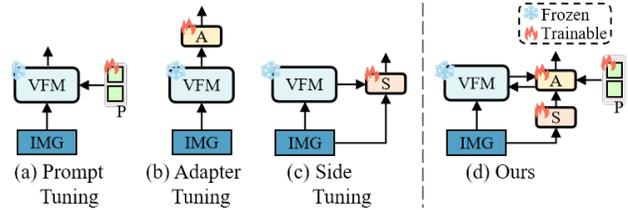

Figure 1. **Previous paradigm vs. our paradigm**. IMG: Input images, P: Prompts, A: Adapter Layers, S: Side Network. Our method builds on three widely-adopted PEFT paradigms. Its key idea involves learnable modules that dynamically interact with the VFM to fully activate and assimilate valuable knowledge, forming world-aware features.

## Introduction

Monocular Depth Estimation (MDE) is the task of predicting per-pixel depth from a single RGB image. As a fundamental yet challenging problem in computer vision, MDE finds important applications in robotics (Luo, Yang and Yuille 2021), autonomous driving (Wang, Pang and Lin 2022), virtual reality(Gerig et al. 2018), and embodied AI (Wang et al. 2024). However, MDE is inherently an ill-posed problem due to scale ambiguity - a single 2D image can correspond to infinitely many 3D scenes.

This ambiguity can only be mitigated through the incorporation of prior knowledge. Previous methods often impose geometric constraints such as planar priors or explicitly discretize the continuous depth range (Fu et al. 2018, Bhat, Alhashim and Wonka 2021, Yang et al. 2023, Bhat, Alhashim and Wonka 2022). While these handcrafted priors partially address the scale ambiguity of MDE, they inevitably limit model expressiveness as they cannot capture the diverse depth patterns present in real-world scenes (Piccinelli, Sakaridis and Yu 2023).

The emergence of vision foundation models (VFMs) has revolutionized the methodology for many computer vision tasks (Caron et al. 2021, Radford et al. 2021, Oquab et al. 2024). Trained via self-supervision on large-scale datasets, VFMs develop comprehensive scene understanding and extract task-agnostic representations, making them suitable for diverse downstream tasks. For MDE, VFMs leverage their powerful world knowledge to extract critical representations that help locate the solution space and produce more physically plausible predictions under real-world constraints (Ke et al. 2024, Yang et al. 2024a, Yang et al. 2024b).

Recent studies have applied VFMs to MDE with promising results, primarily via full-parameter fine-tuning (FPFT). However, FPFT with limited data presents several challenges: the high degree of learnable freedom increase overfitting risks; excessive modification of core pre-trained parameters may erase general prior; and insufficient task-specific data can hinder learning of new patterns while degrading existing representations (He et al. 2023, Xin et al. 2024).

To lower the barrier of adapting VFMs and broaden their applicability, we propose a novel approach called WEDepth, based on parameter-efficient-fine-tuning (PEFT) techniques (e.g., prompt, adapter and side tuning, Fig. 1) that effectively transfers their valuable world-understanding priors to MDE tasks while keeping the base model structure and weights unchanged. As shown in Fig. 2, our method employs a dual-branch design: an encoder-decoder branch serves as the predictor for per-pixel depth estimation; and a VFM branch acts as an enhancer, providing general world knowledge.

The key innovation is our proposed Partition-Enhance-Inject (PEI) module that facilitates information exchange between the predictor and enhancer. It operates through three phases: (1) **Partition**: Multi-scale feature maps from the encoder of predictor are processed to adaptively extract critical patterns; (2) **Enhance**: Extracted patterns are combined with image tokens and fed into the VFM's transformer blocks, enabling mutual interaction via pretrained attention mechanisms. Pattern tokens aggregate general image representations, while image tokens receive MDE-adaptive conditional prompts; (3) **Inject**: Enhanced tokens are reintegrated into the encoder, where pattern tokens are adaptively transformed into continuous pixel feature space and image tokens are spatially aligned with pixel feature via interpolation. Finally, the PEI-augmented multi-scale features are processed by the decoder for metric depth estimation.

The main contributions are summarized as follows:

(1) We propose a novel approach to adapt VFMs for MDE. Our method employs the VFM as a feature enhancer while preserving its original structure and weights, effectively leveraging the model's robust world priors to augment task-relevant features. This work establishes a new paradigm for transferring VFM to downstream vision tasks.

(2) We introduce a PEI module to facilitate information exchange between the VFM and the depth predictor. This module effectively activates and assimilates the world knowledge embedded in the VFM, endowing the MDE-relevant features with world-aware capabilities. The PEI significantly mitigates the ill-posed nature of MDE tasks.

(3) Extensive experiments demonstrate that our method achieves competitive performance compared to current state-of-the-art (SOTA) methods on both NYU Depth v2 and KITTI datasets. Notably, current SOTA methods rely on diffusion models requiring multiple forward passes, or fine-tuning relative depth prediction models that necessitate large-scale pretraining. In contrast, our method only requires training on a single dataset and performs inference via a single forward pass, reducing both training and inference costs. Furthermore, our method exhibits superior generalization across multiple zero-shot transfer tasks.

# Related Work

**Monocular Depth Estimation.**
Deep learning has greatly advanced MDE, beginning with Eigen et al.'s multi-scale fusion network (Eigen, Puhrsch and Fergus 2014). Subsequent researches incorporates various geometric priors to enhance MDE, including: normal constraints that enforce consistency between surface normals derived from estimated and ground truth depths (Qi et al. 2018, Long et al. 2021); the local planar assumption, which employs multi-scale local planar guidance layers to establish direct relationships between internal features and predicted depth (Lee et al. 2019); planar parallax geometry that decomposes inter-frame correspondences into planar homography and residual disparity for more stable reconstruction (Xing et al. 2022, Yuan et al. 2023); and the planar ground prior, which utilizes ground plane references to improve accuracy and generalizability (Yang et al. 2023). In contrast to these geometry-based approaches, our method fully leverages the world knowledge embedded in VFMs to comprehensively capture scene information.

**Vision Foundation Model.**
VFMs are typically trained through self-supervised learning on internet-scale data, acquiring task-agnostic general representations with advanced visual understanding. These representations can be directly applied to downstream tasks with minimal fine-tuning effort (Bommasani et al. 2021). Full-parameter fine-tuning is effective when sufficient task-specific data and computational resources are available with limited distribution shift; otherwise, PEFT methods are preferred.

Among prevalent PEFT techniques: (1) Prompt Tuning introduces learnable prompt tokens at the input or intermediate layers to guide the model toward specific outputs (Zhu et al. 2023, Bahng et al. 2022) (2) Adapter Tuning inserts small bottleneck adapter modules (such as fully-connected layers) between Transformer layers, updating only the adapter parameters (Zhang et al. 2021, Pantazis et al. 2022);. (3) Side Tuning use a smaller, detached side network running alongside the pretrained models (Zhang et al. 2020, Chen et al. 2022).

Unlike conventional approaches that directly adapt VLMs to downstream tasks, our method leverages the VLM as an auxiliary feature enhancer, preserving the model's generalizability while augmenting task-specific features.

**VFMs used for MDE.**
Recent works have demonstrated that VLMs can significantly improve the accuracy and generalization of MDE. VPD employs Stable Diffusion model as image feature extractor augmented with text inputs (Zhao et al. 2023). DepthGen (Saxena et al. 2023) and DDP (Ji et al. 2023) implement noise-to-depth paradigms. ECoDepth (Patni, Agarwal and Arora 2024) incorporates a diffusion backbone conditioned on ViT embeddings. Depth Anything (Yang et al. 2024a, Yang et al. 2024b) fine-tunes DINO v2 (Oquab et al. 2024) using both labeled and unlabeled data to create a MDE model with strong zero-shot transfer. VGGT (Wang et al. 2025a) and MoGe (Wang et al. 2025b) construct models based on DINOv2, demonstrating robust monocular 3D geometry estimation to various 3D perception tasks.

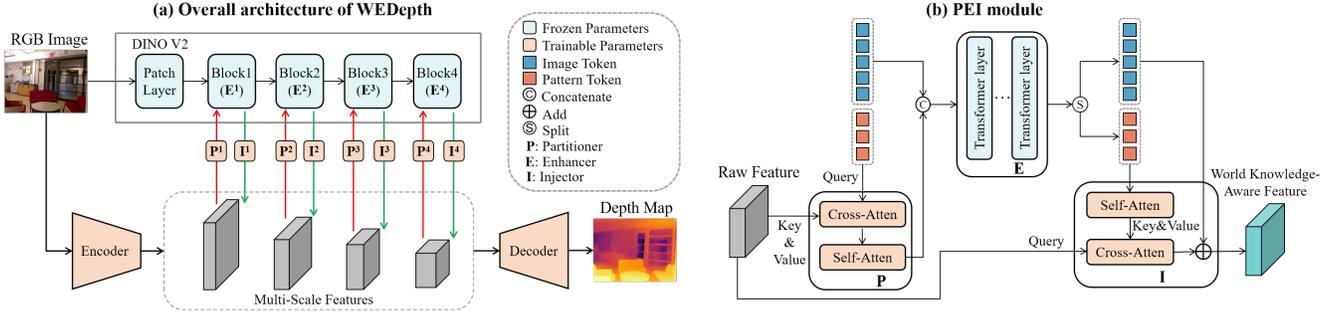

Figure 2. **An overview of our proposed model.** The multi-scale feature maps extracted by the encoder interact with the generic features from the frozen DINOv2 through our proposed **P**artition-**E**nhance-**I**nject (PEI) mechanism, enabling adaptive aggregation of world knowledge. In the PEI mechanism, the pattern tokens partitioned from the feature maps, along with the image tokens, are fed into the transformer blocks of DINOv2. Through the processing of the pre-trained attention mechanism, both types of tokens are enhanced. Finally, they are reinjected into the raw feature map, yielding world-knowledge-aware features.

While pre-trained diffusion models excel at image generation tasks, their requirement for multiple iterative denoising steps results in high computational costs, making them impractical for real-time processing. In contrast, our method employs DINOv2, a plain ViT-based architecture (Dosovitskiy et al. 2020), which requires only a single forward pass and exhibits robust cross-domain understanding for both image- and pixel-level tasks.

## Approach

### Overall architecture

The overall architecture of WEDepth is illustrated in Fig. 1, which can be divided into two branches: a depth predictor with an encoder-decoder structure and a feature enhancer composed of DINOv2. The PEI mechanism establishes information interaction between them for feature enhancement. First, the input image is fed into the encoder to extract multi-scale features. Simultaneously, the input image is processed through the patch layer of the enhancer to obtain image tokens. Subsequently, the PEI mechanism is employed to extract key patterns embedded in features at each scale. These pattern tokens are then combined with the image tokens and input into the transformer block of the enhancer for interactive processing, yielding enhanced pattern tokens and image tokens. Following this, the two types of tokens are fused with the corresponding scale features in encoder. Finally, these enhanced features are fed into the decoder for depth estimation.

### Image embedding

We employ a ResNet or Swin-Transformer backbone as the encoder to map the image into features of four distinct scales ($\{F^l\}_{l=1}^4$). These feature maps exhibit progressively reduced resolutions (1/4, 1/8, 1/16, and 1/32 of the original image) while their channel dimensions incrementally expand.

We utilize DINOv2-Large as enhancer, which adopts a plain ViT architecture comprising a patch embedding layer Patch(·) followed by a series of Transformer layers. This enhancer projects the image into 1024-dimensional tokens, reducing the feature resolution to 1/14 of the original image. To align with the four-scale features from the encoder, the Transformer layers are partitioned into four blocks $\{Block^l(\cdot)\}_{l=1}^4$, thereby enabling feature enhancement across the four scales via the PEI module.

### PEI module

The PEI module facilitates the adaptation of powerful world priors from VLMs to MDE task. It operates sequentially at each feature in $\{F^l\}_{l=1}^4$ through three key stages: (1) Pattern Partitioning, which identifies and extracts essential patterns $P^l$ from the input features $F^l$; (2) Feature Enhancement, where the pattern tokens $P^l$ combined with image tokens $T^l$ are refined through $Block^l$ to produce enhanced features; and (3) Knowledge Injection, which adaptively integrates the knowledge in the enhanced features back into the $F^l$ to generate the world-aware feature $\tilde{F}^l$.

**Pattern Partitioning.** For each feature map $F^l$, we introduce a set of learnable identity embeddings (Carion et al. 2020) $I^l \in \mathbb{R}^{N \times C}$ to define pattern identities ($N$=50 in this work). They are combined with semantic embeddings $S^l \in \mathbb{R}^{N \times C}$ ($S$ is zero-initialized for $l$=1, otherwise derived from the output of $Block^{l-1}(\cdot)$) to form prior pattern queries $Pri^l \in \mathbb{R}^{N \times C} = I^l + S^l$. Through a cross-attention layer where $Pri^l$ serves as queries and $F^l$ as keys/values and a self-attention layer the module output conditional pattern embeddings $P^l$ that aggregate contextual information from $F^l$. The processing is defined as:

$$P^l = \text{Liner}\left(\text{SelfAttn}\left(\text{CrossAttn}(Pri^l, F^l)\right)\right) \quad (1)$$

Where Liner(·) is linear projection layer that aligns embeddings to the same channel dimension as DINOv2 tokens.

| Method | AbsRel↓ | RMSE↓ | log10↓ | SqRel↓ | δ1↑ | δ2↑ | δ3↑ |
|---|---|---|---|---|---|---|---|
| Eigen et al. (Eigen, Puhrsch and Fergus 2014) | 0.158 | 0.641 | - | - | 0.769 | 0.95 | 0.988 |
| DORN (Fu et al. 2018) | 0.115 | 0.509 | 0.051 | - | 0.828 | 0.965 | 0.992 |
| SharpNet (Ramamonjisoa and Lepetit 2019) | 0.139 | 0.502 | 0.047 | - | 0.836 | 0.966 | 0.993 |
| BTS (Lee et al. 2019) | 0.11 | 0.392 | 0.047 | 0.066 | 0.885 | 0.978 | 0.994 |
| AdaBins (Bhat, Alhashim and Wonka 2021) | 0.103 | 0.364 | 0.044 | - | 0.903 | 0.984 | 0.997 |
| DPT (Ranftl, Bochkovskiy and Koltun 2021) | 0.11 | 0.357 | 0.045 | - | 0.904 | 0.988 | 0.998 |
| P3Depth (Patil et al. 2022) | 0.104 | 0.356 | 0.043 | - | 0.898 | 0.981 | 0.996 |
| NeWCRFs (Yuan et al. 2022) | 0.095 | 0.334 | 0.041 | 0.045 | 0.922 | 0.992 | 0.998 |
| SwinV2 (Liu et al. 2022) | 0.112 | 0.381 | 0.051 | - | 0.886 | 0.984 | 0.997 |
| Localbins (Bhat, Alhashim and Wonka 2022) | 0.099 | 0.357 | 0.042 | - | 0.907 | 0.987 | 0.998 |
| Jun et al. (Jun et al. 2022) | 0.098 | 0.355 | 0.042 | - | 0.913 | 0.987 | 0.998 |
| PixelFormer (Agarwal and Arora 2023) | 0.09 | 0.322 | 0.039 | 0.043 | 0.929 | 0.991 | 0.998 |
| iDisc (Piccinelli, Sakaridis and Yu 2023) | 0.086 | 0.313 | 0.042 | | 0.940 | 0.993 | 0.999 |
| MIM (Xie et al. 2023) | 0.083 | 0.287 | 0.035 | - | 0.949 | 0.994 | 0.999 |
| AiT (Ning et al. 2023) | 0.076 | 0.275 | 0.033 | - | 0.954 | 0.994 | 0.999 |
| DDP + (Ji et al. 2023) | 0.094 | 0.329 | 0.04 | - | 0.921 | 0.99 | 0.998 |
| VPD + (Zhao et al. 2023) | 0.069 | 0.254 | 0.030 | 0.027 | 0.964 | 0.995 | 0.999 |
| EcoDepth + (Patni, Agarwal and Arora 2024) | <u>0.059</u> | <u>0.218</u> | <u>0.026</u> | **0.013** | <u>0.978</u> | <u>0.997</u> | 0.999 |
| ZoeDepth * (Bhat et al. 2023) | 0.075 | 0.27 | 0.032 | 0.030 | 0.955 | 0.995 | 0.999 |
| Depth Anything * (Yang et al. 2024a) | **0.056** | **0.206** | **0.024** | | **0.984** | **0.998** | **1.000** |
| **Ours (ResNet-101)** | 0.068 | 0.243 | 0.029 | 0.023 | 0.974 | <u>0.997</u> | 0.999 |
| **Ours (Swin-Small)** | 0.072 | 0.251 | 0.031 | 0.025 | 0.972 | **0.998** | 0.999 |
| **Ours (Swin-Base)** | 0.066 | 0.237 | 0.028 | <u>0.022</u> | 0.976 | **0.998** | **1.000** |

Table 1. **Results on Indoor NYU Depth v2 Dataset**. Best results are in bold, second best are underlined. Symbol (+) indicates that the method is based on a diffusion model and requires multiple feedforward processing steps. Symbol (*) indicates that the method is pre-trained on relative depth.

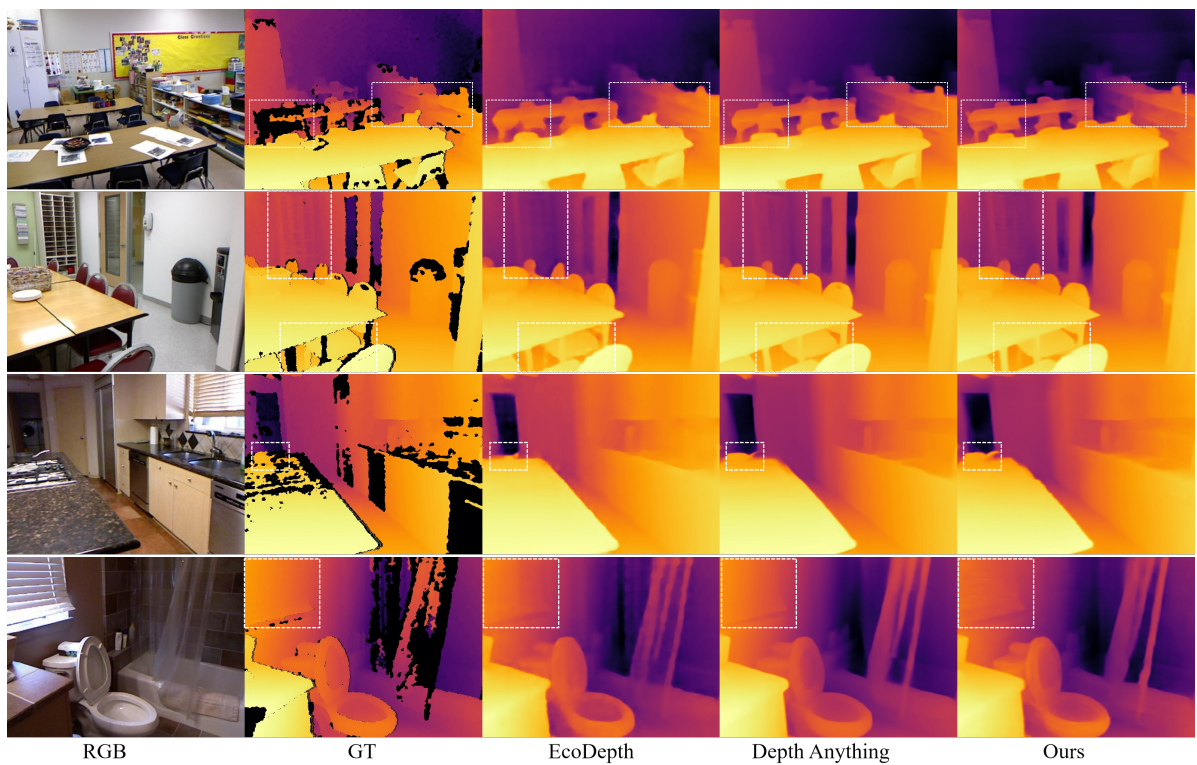

Figure 2. **Visual Comparison on NYU Depth v2 Indoor Dataset**.

| Method | AbsRel↓ | SqRel↓ | RMSElog↓ | RMSE↓ | δ1↑ | δ2↑ | δ3↑ |
|---|---|---|---|---|---|---|---|
| Eigenetal. (Eigen, Puhrsch and Fergus 2014) | 0.203 | 1.517 | 0.282 | 6.307 | 0.702 | 0.898 | 0.967 |
| DORN (Fu et al. 2018) | 0.072 | 0.307 | 0.12 | 2.727 | 0.932 | 0.984 | 0.994 |
| BTS (Lee et al. 2019) | 0.059 | 0.241 | 0.096 | 2.756 | 0.956 | 0.993 | 0.998 |
| AdaBins (Bhat, Alhashim and Wonka 2021) | 0.067 | 0.19 | 0.088 | 2.96 | 0.949 | 0.992 | 0.998 |
| DPT (Ranftl, Bochkovskiy and Koltun 2021) | 0.06 | - | 0.092 | 2.573 | 0.959 | 0.995 | 0.996 |
| P3Depth (Patil et al. 2022) | 0.071 | 0.27 | 0.103 | 2.842 | 0.953 | 0.993 | 0.998 |
| NeWCRFs (Yuan et al. 2022) | 0.052 | 0.155 | 0.079 | 2.129 | 0.974 | 0.997 | 0.999 |
| PixelFormer (Agarwal and Arora 2023) | 0.051 | 0.149 | 0.077 | 2.081 | 0.976 | 0.997 | 0.999 |
| iDisc (Piccinelli, Sakaridis and Yu 2023) | 0.050 | 0.145 | 0.077 | 2.067 | 0.977 | 0.977 | - |
| IEBins (Shao et al. 2023) | 0.050 | 0.142 | 0.075 | 2.011 | 0.978 | 0.998 | 0.999 |
| MIM (Xie et al. 2023) | 0.050 | 0.139 | 0.075 | 1.966 | 0.977 | 0.998 | 1.000 |
| GEDepth (Yang et al. 2023) | 0.048 | 0.142 | 0.076 | 2.044 | 0.976 | 0.997 | 0.999 |
| DDP + (Ji et al. 2023) | 0.050 | 0.148 | 0.076 | 2.072 | 0.975 | 0.997 | 0.999 |
| EcoDepth + (Patni, Agarwal and Arora 2024) | 0.048 | 0.139 | 0.074 | 2.039 | 0.979 | 0.998 | 0.999 |
| ZoeDepth * (Bhat et al. 2023) | 0.054 | 0.189 | 0.083 | 2.44 | 0.97 | 0.996 | 0.999 |
| Depth Anything * (Yang et al. 2024a) | **0.046** | - | **0.069** | **1.896** | **0.982** | **0.998** | **1.000** |
| Ours (ResNet-101) | 0.054 | 0.150 | 0.078 | 2.085 | 0.977 | **0.998** | **1.000** |
| Ours (Swin-Small) | 0.050 | 0.138 | 0.074 | 1.964 | 0.980 | **0.998** | **1.000** |
| Ours (Swin-Base) | 0.048 | **0.127** | 0.072 | 1.936 | **0.982** | **0.998** | **1.000** |

Table2. **Performance on the Outdoor KITTI Dataset**. Please refer to the caption of Table. 1 for notation details

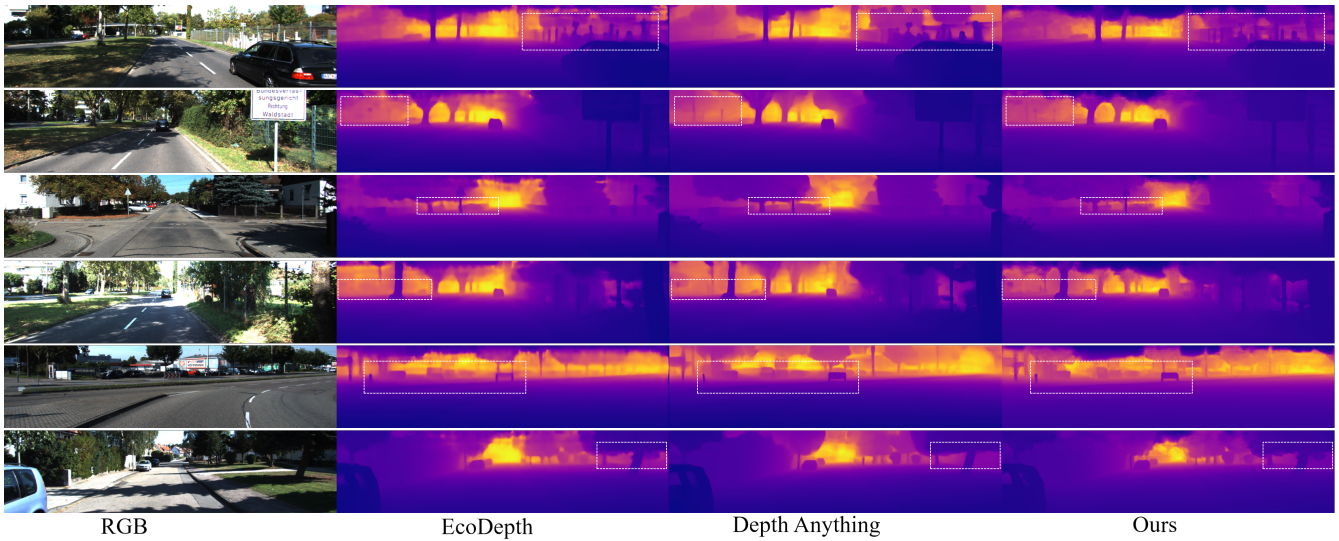

Figure 5. **Visual Comparison on KITTI Outdoor Dataset**.

**Feature Enhancement.** $P^l$ is concatenated with image tokens $T^l$ and processed by the $Block(\cdot)^l$ ($T^l$ are output from Patch($\cdot$) when $l$=1, otherwise derived from the output of $Block^{l-1}(\cdot)$). We implement an attention mask $\mathcal{M}$ restricting image tokens $T^l$ to freely interact while preventing inter-pattern communication in $P^l$, thereby maintaining pattern discreteness. The processing is defined as:

$$\bar{P}^l, T^{l+1} = \text{Block}^l(P^l \oplus T^l, \mathcal{M}) \quad (2)$$

where $\bar{P}^l$ is enhanced pattern tokens. Notably, the $\bar{P}^l$ output from $Block^l$ should combined with $I^{l+1}$ to form $Pri^{l+1}$:

$$Pri^{l+1} = \bar{P}^l + I^{l+1} \quad (3)$$

which is then input into Eq (1) to calculate the $P^{l+1}$ of the $Block^{l+1}$.

This stage enhances features through two mechanisms: (1) Enriches $P^l$ with global semantic information from $T^l$, enhancing world knowledge; (2) $P^l$ can be regarded as a conditional prompt for $T^l$, transforming generic representations into depth-aware semantics. The shared pretrained transformer block jointly processes $P^l$ and $T^l$, enabling bidirectional enhancement that yields more diverse priors better suited for MDE.

|  | SUN RGB-D | | | iBims-1 Benchmark | | |
|---|---|---|---|---|---|---|
| Method | δ1↑ | AbsRel↓ | RMSE↓ | δ1↑ | AbsRel↓ | RMSE↓ |
| BTS (Lee et al. 2019) | 0.74 | 0.172 | 0.515 | 0.538 | 0.231 | 0.919 |
| AdaBins (Bhat, Alhashim and Wonka 2021) | 0.771 | 0.159 | 0.476 | 0.555 | 0.212 | 0.901 |
| LocalBins (Bhat, Alhashim and Wonka 2022) | 0.777 | 0.156 | 0.47 | 0.558 | 0.211 | 0.880 |
| NeWCRFs (Yuan et al. 2022) | 0.777 | 0.156 | 0.47 | 0.548 | 0.206 | 0.861 |
| VPD (Zhao et al. 2023) | 0.861 | 0.121 | 0.355 | 0.627 | 0.187 | 0.767 |
| ZoeDepth (Bhat et al. 2023) | 0.864 | 0.119 | 0.346 | 0.658 | 0.169 | 0.711 |
| EcoDepth (Patni, Agarwal and Arora 2024) | <u>0.885</u> | <u>0.112</u> | <u>0.319</u> | <u>0.688</u> | <u>0.163</u> | **0.664** |
| **Ours** | **0.902** | **0.105** | **0.309** | **0.738** | **0.152** | <u>0.666</u> |

Table3. **Zero-shot testing of models trained on NYU**. Evaluation depth is capped at 8m for SUN RGB-D, 10m for iBims. Best results are in bold, second best are underlined.

|  | Argoverse | | | DDAD | | |
|---|---|---|---|---|---|---|
| Method | δ1↑ | RMSE↓ | SIlog↓ | δ1↑ | RMSE↓ | SIlog↓ |
| BTS (Lee et al. 2019) | 0.307 | 15.98 | 0.518 | 0.399 | 16.19 | 0.405 |
| AdaBins (Bhat, Alhashim and Wonka 2021) | 0.383 | 17.07 | 0.523 | 0.282 | 18.36 | 0.507 |
| P3Depth (Patil et al. 2022) | 0.277 | 17.97 | 0.441 | 0.397 | 17.83 | 0.390 |
| NeWCRF (Yuan et al. 2022) | 0.311 | 15.75 | 0.468 | 0.343 | 16.76 | 0.442 |
| iDisc (Piccinelli, Sakaridis and Yu 2023) | <u>0.560</u> | <u>12.18</u> | **0.334** | <u>0.350</u> | <u>14.26</u> | **0.294** |
| **Ours** | **0.594** | **12.11** | <u>0.398</u> | **0.442** | **14.21** | <u>0.355</u> |

Table4. **Zero-shot testing of models trained on KITTI**. The splits of Argoverse and DDAD is according to (Piccinelli, Sakaridis and Yu 2023). Evaluation depth is capped at 150m for them. Best results are in bold, second best are underlined.

**Knowledge Injection**. After completing the feature enhancement, we obtain the $\{\bar{P}^l\}_{l=1}^4$. Each $\bar{P}^l$ is projected to match channel dimension of $F^l$ and proposed by a self-attention layer, then interact with $F^l$ via a cross-attention layer, where $F^l$ server as queries and $\bar{P}^l$ server as keys/values:

$$\bar{F}^l = F^l + \text{CrossAttn}\left(F^l, \text{SelfAttn}(\text{Liner}(\bar{P}^l))\right) \quad (4)$$

Where $\bar{F}^l$ represents features after integrating enhanced patterns. This selectively reintegrates critical patterns into the original feature, enabling pixel-wise feature enhancement through adaptive aggregation of world knowledge.

The enhancer achieves prompt-learning effects through incorporating $P^l$. Under the guidance of $P^l$, the generic representations gradually adapt to the MDE domain while providing global context supplementation. Therefore, we extract the image tokens $T^f$ from the last ViT layer as they contain sufficient semantics to describe hierarchical features(Li et al. 2022) (our analysis also shows that tokens from the last layer are more effective than those from the intermediate layer.), resize it via bilinear interpolation to multiple resolutions, and perform pixel-wise addition with corresponding enhanced features $\bar{F}^l$:

$$\tilde{F}^l = \text{Resize}^l(T^f) + \bar{F}^l \quad (5)$$

where $\tilde{F}^l$ represents the final world-aware feature. This dual-branch fusion strategy has demonstrated effectiveness in dense prediction tasks(Xia et al. 2024).

## Decoder

The $\tilde{F}^l$ at each scale are processed using multi-scale deformable attention (MSDA) decoder. The decoded feature maps at different scales are individually employed for depth estimation, and the final depth prediction is the average of the outputs from all scales.

## Experiments

### Implementation Details

**Datasets.** We employ NYU Depth V2 (indoor) (Silberman et al. 2012) and KITTI (outdoor) (Geiger et al. 2013) as primary training datasets. NYU Depth V2 contains 464 scenes with 24,000+ training and 654 test RGB-depth pairs (640×480), captured by structured light sensors. KITTI provides 24,000+ stereo-LiDAR pairs (1241×376) from vehicle-mounted cameras. Dataset is split via the Eigen protocol (Eigen, Puhrsch and Fergus 2014), yielding 23,158 training and 652 test samples.

For zero-shot evaluation, we assess models trained on NYU Depth V2 or KITTI by directly testing them on complementary indoor or outdoor datasets (Sun-RGBD (Song, Lichtenberg and Xiao 2015), iBims-1(Koch et al. 2018) Argoverse 1.1(Chang et al. 2019), and DDAD (Guizilini et al. 2020)) without any additional fine-tuning.

**Training.** We implement our model using PyTorch. The model is optimized with the AdamW (Loshchilov and Hutter 2017) ($\beta_1 = 0.9$, $\beta_2 = 0.999$, weight decay=0.01), trained for 20 epochs (batch size = 8, initial learning rate = $2\times10^{-3}$) using linear warmup (3 epochs) followed by cosine decay. The optimization process is guided by Scale-Invariant loss (Eigen, Puhrsch and Fergus 2014). Mixed-precision training is adopted to accelerate the training process. All encoder backbones are initialized with ImageNet pretrained weights. Training completes in approximately 14 minutes per epoch on four NVIDIA A100 GPUs. More experimental details see supplementary materials.

|   | P | E | I | | AbsRel↓ | RMSE↓ | δ1↑ |
|   |   |   | PT | IT |   |   |   |
|---|---|---|---|---|---|---|---|
| 1 | ✗ | ✗ | ✗ | ✗ | 0.110 | 0.369 | 0.892 |
| 2 | ✗ | ✓ | ✓ | ✗ | 0.084 | 0.300 | 0.946 |
| 3 | ✓ | ✗ | ✓ | ✗ | 0.108 | 0.367 | 0.897 |
| 4 | ✗ | ✓ | ✗ | ✓ | 0.072 | 0.254 | 0.969 |
| 5 | ✓ | ✓ | ✓ | ✗ | 0.081 | 0.297 | 0.946 |
| 6 | ✓ | ✓ | ✗ | ✓ | 0.072 | 0.250 | 0.969 |
| 7 | ✗ | ✓ | ✓ | ✓ | 0.067 | 0.244 | 0.072 |
| 8 | ✓ | ✓ | ✓ | ✓ | 0.066 | 0.237 | 0.976 |

Table 5 **Ablation of WEDepth**. P: Partitioning, E: Enhancement, I: Injection, PT: Pattern Tokens, IT: Image Tokens. When the Partitioning component is removed, the pattern token is just a set of learnable parameters that do not interact with the features in the encoder (rows 2, 4, and 7).

## Comparison on NYU Depth v2

Table 1 presents a comparison between our method and SOTA methods on the NYU depth V2. Our method maintains competitive performance even when compared against recent diffusion-based approaches and methods utilizing relative depth pre-training strategies. Among all evaluated methods, EcoDepth, Depth Anything, and our method consistently achieve superior performance across all metrics, demonstrating significant margins over rest competing methods. Qualitative comparisons on NYU Depth v2 (Fig. 3) reveal our method's superior in reconstructing geometric details of both diminutive objects and distant small-scale structures, such as table legs, internal shelves of cabinets, and pot handles. Notably, our approach can robustly reconstruct fine-grained structures even in the absence of corresponding ground truth information, as evidenced by the precisely rendered light gaps in blackout curtains (row 5).

## Comparison on KITTI

As shown in Table 2, our method also achieves competitive performance on the outdoor KITTI dataset. While existing SOTA approaches demonstrate saturated performance gains in this domain, our method maintains comparable accuracy to Depth Anything while producing more structurally coherent depth maps. Qualitative results (Fig.4) validate our method's superior preservation of fine-grained edges and details for distant objects, such as highway guardrails, tree trunks, and occluded vehicles.

## Generalization and Zero Shot Transfer

Table 3 presents the zero-shot evaluation results of the NYU Depth v2 model on the indoor datasets SUN-RGBD and DI-ODE, demonstrating our model's strong generalization to unseen datasets. Furthermore, we evaluated the zero-shot performance of the KITTI model on Argoverse and DDAD datasets, shown in Table 3. Despite the significant disparity in depth ranges between KITTI (0-80m) and Argoverse/DDAD (0-150m), our model maintains robust performance, which further suggest its generalization across different depth scales and scenarios.

## Ablation Analysis

Table 5 presents a comprehensive ablation study evaluating the contribution of each component in our method. Using an encoder-decoder predictor as the baseline, we incrementally integrate components to assess their impact.

**Feature Enhancement Analysis**. Significant performance gains are observed when integrating features from the DINOv2 through pattern token fusion (Row 2 vs. Row 1) or image token (Row 4 vs. Row 1). These results substantiate that the generalized representations and prior knowledge embedded in the VFMs are essential for advancing monocular 3D geometric comprehension.

**Feature Injection Analysis**. The fusion of image tokens demonstrates more pronounced improvements compared to pattern token fusion (Row 4 vs. Row 2 and Row 6 vs. Row 5). Notably, concurrent fusion of both token types yields further improvements, suggesting their complementary nature and synergistic interaction when combined.

**Pattern Partitioning Analysis**. The pattern partition module provides only marginal RMSE improvements when independently fusing image tokens (Row 5 vs. Row 2) or pattern tokens (Row 6 vs. Row 4). However, when fusing both token types, the partition module delivers substantial performance gains (Row 8 vs. Row 7). This indicates that the pattern partition module effectively enhances the representational capacity of both type tokens, enabling more discriminative feature capture.

## Conclusion

In this paper, we present WEDepth, a novel approach for MDE. Our key idea lies in effectively leveraging the implicit world prior knowledge embedded in VFMs, rather than relying on explicit geometric priors. We realize this through our proposed PEI mechanism, which employs VFMs to enhance features in trainable modules via a Partition-Enhance-Inject pipeline. This mechanism comprehensively activates and assimilates the prior knowledge from VFMs, endowing the extracted task-specific features with robust world understanding capabilities, thereby mitigating the ill-posed nature of the MDE task. Extensive experiments have been conducted to demonstrate the effectiveness of our approach.